\documentclass{article}
\usepackage{iclr2024_conference, times}
\usepackage[utf8]{inputenc} % allow utf-8 input
\usepackage[T1]{fontenc}    % use 8-bit T1 fonts
\usepackage{nicefrac}       % compact symbols for 1/2, etc.
\usepackage{microtype}
\usepackage{amsmath}
\usepackage{amssymb}
\usepackage{amsfonts}       % blackboard math symbols
\usepackage{graphicx}

\usepackage{caption}
\usepackage{subcaption}
\usepackage{tabularx}
\usepackage{multirow}
\usepackage{booktabs}
\usepackage{appendix}
\usepackage{lscape}
\usepackage{url}

\title{Generalizable Temperature Nowcasting with Physics-Constrained RNNs for Predictive Maintenance of Wind Turbine Components}

\author{Johannes Exenberger \\
TU Graz\\
\texttt{\small{johannes.exenberger@tugraz.at}} \\
\And
Matteo Di Salvo \\
Sirius Energy Automation \\
\texttt{\small{disalvo@sirius-ea.com}}
\And
Thomas Hirsch \\
TU Graz\\
\And
Franz Wotawa \\
TU Graz\\
\And
Gerald Schweiger \\
TU Graz\\
}

\iclrfinalcopy % Uncomment for camera-ready version, but NOT for submission.

\begin{document}

\maketitle

\begin{abstract}
Machine learning plays an important role in the operation of current wind energy production systems. One central application is predictive maintenance to increase efficiency and lower electricity costs by reducing downtimes. Integrating physics-based knowledge in neural networks to enforce their physical plausibilty is a promising method to improve current approaches, but incomplete system information often impedes their application in real world scenarios. We describe a simple and efficient way for physics-constrained deep learning-based predictive maintenance for wind turbine gearbox bearings with partial system knowledge. The approach is based on temperature nowcasting constrained by physics, where unknown system coefficients are treated as learnable neural network parameters. Results show improved generalization performance to unseen environments compared to a baseline neural network, which is especially important in low data scenarios often encountered in real-world applications.
\end{abstract}

\section{Introduction}

Machine learning (ML) methods are a crucial part of the wind energy sector \citep{maruganSurveyArtificialNeural2018} and help improving operational efficiency of wind power plants to make electricity prices competitive with currently dominant fossil-based energy sources \citep{shafieeMaintenanceOptimizationInspection2019a}. Predictive maintenance is applied to prevent component failures and increase maintenance efficiency.
Models of component deterioration processes estimate when maintenance should be performed before damages occur, resulting in shorter downtimes. For large wind farms and those with restricted access such as offshore plants, maintenance is costly and time consuming \citep{carrollAvailabilityOperationMaintenance2017}. Optimized maintenance scheduling is thus highly desired, with estimated reductions in the Levelized Cost of Electricity of up to 30\% for onshore wind farms \citep{costaNewTendenciesWind2021}. Especially gearbox faults, mainly caused by bearing failure, can lead to long downtimes \citep{hahnReliabilityWindTurbines2007, rederWindTurbineFailures2016, wangIntegratedFaultDiagnosis2020}. 
The core of predictive maintenance is a surrogate model - either data-driven or physics-based - that describes the system in normal conditions. Deviations between this baseline and current measurements triggers an alarm. The system dynamics are often only partially known, making physics-based models difficult to adapt to each scenario when system-specific measurements are not available, while data driven methods don't guarantee physical plausibility. The integration of physical laws in ML models known as physics-informed ML (PIML) \citep{karniadakisPhysicsinformedMachineLearning2021} allows to combine both approaches.
In this paper, we describe a neural network-based approach for predictive maintenance of a wind turbine generator (WTG) gearbox bearing constrained by physics without the need for precise measurements of component coefficients, which are treated as trainable parameters\footnote{The code is available at: \url{github.com/jxnb/pcrnn-wtg}}\footnote{The dataset is available upon request to the authors.}. 
By performing nowcasting, i.e. predicting the current state rather than a future state, (1) the inference process does not depend on wind speed forecasting which introduces additional uncertainties, and (2) external constraints such as for curtailment periods do not affect the model.
This approach shows competitive prediction performance and improved generalization capabilities in comparison to a physics-based model and a baseline neural network. The approach is not limited to WTGs, but applicable to a wide range of systems where partially known physics-based knowledge can be integrated in the modelling process.

\section{Model description} \label{sec:model}

We apply a common procedure of ML-based predictive maintenance: a surrogate model trained on observations represents the expected system behavior in normal conditions. Over the system's lifetime, the deviations between estimated and true behavior are expected to increase with growing wear of the components, providing information about the condition of the system and potential need for maintenance. 
Our approach aims to circumvent common problems in ML-based predictive maintenance for WTGs. First, data on system faults is usually scarce - WTGs have availability rates in the order of 98\% \citep{yangWindTurbineCondition2013}; component faults are rare and occur mostly at the end of a component lifetime period. This limits the applicability of fault detection methods. 
Second, ML models leveraging physics often require knowledge of specific component coefficients \citep[e.g.][]{yucesanHybridPhysicsinformedNeural2022}. This information is often not available or unreliable, as components like the gearbox vary even between WTGs of the same model. Treating these unknown coefficients as trainable parameters allows to incorporate physics into the model when they are not or only partially known, simplifying the adaptation of this approach to other applications. Third, wind speed is the main driver of bearing temperatures in WTGs, making component temperature prediction mainly a wind speed forecasting problem. Local wind speeds are very difficult to predict also for short time horizons of several minutes due to complex local turbulence patterns. As we focus on nowcasting, i.e. the estimation of the current bearing temperature state, the current rotor speed as the main driver is known. This allows us to avoid the large factor of uncertainty associated with wind speed forecasts and increases prediction performance during periods where wind speed and power production are uncorrelated, such as curtailment for wind farms due to, for example, grid loads constraints or bird protection.

\subsection{Physics model}

The physics-based description of a change in bearing temperature $T^b$ between two time steps is given by the following heat transfer ODE equation for transient-state, lumped systems based on \cite{zhangComparisonDatadrivenModelbased2014, cambronBearingTemperatureMonitoring2017}:
\begin{align}
\label{eq:physics_model}
    C_p \frac{dT^b}{dt} \approx R^{-1} (T^a_{t+1} - T^b_t) + \mu \omega_{t+1} + \alpha P_{t+1}
\end{align}
where $C_p$ [$\text{W K}^{-1}$] is the nacelle heat capacity, $R^{-1} (T^a_{t+1} - T^b_t)$ is the system's thermal conductivity with the resistance coefficient $R$ [$\text{K W}^{-1}$], bearing temperature $T^b_t$ [K] and ambient temperature $T^a_{t+1}$ [K]. $\mu \omega_{t+1}$ is the heat produced by friction in the rotating gearbox shaft with friction coefficient $\mu$ [$\text{Ws}^2 \text{rad}^{-2}$] and rotational frequency $\omega_{t+1}$ [$\text{rad s}^{-1}$]. $\alpha P_{t+1}$ is the amount of energy degradation as heat per unit of power $P$ [KW] produced.
The coefficients $C_P$, $R$, $\mu$ and $\alpha$ are unknown and depend on the complexity of the system and materials design and have to be approximated by the model. Based on Equation \ref{eq:physics_model}, the equation solved by the neural network with trainable network parameters $\lambda_1, \lambda_2, \lambda_3$ can be formulated as:
\begin{align}
\label{eq:dT_neuralnetwork}
    \Delta T^b_{t-1} \approx \lambda_1 (T^a_t - T^b_{t-1}) + \lambda_2 \omega_t + \lambda_3 P_t
\end{align}

\subsection{Physics-constrained recurrent neural network} \label{sec:pcrnn}

\begin{figure}
    \centering
    \includegraphics[width=\textwidth]{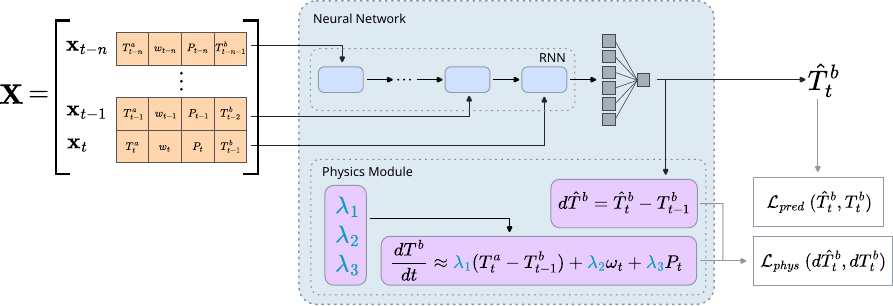}
    \caption{Physics-constrained recurrent neural network (PC-RNN).}
    \label{fig:model}
    \vspace{-1em}
\end{figure}

Given the physics-based model in Equation \ref{eq:physics_model}, the observed state of a WTG at time $t$ is described by the vector $\mathbf{x}_t = (T^a_t$, $\omega_t$, $P_t$, $T^b_{t-1})$. The objective of the neural network is the prediction of the bearing temperature at current time $t$, $\hat{T}^b_t$. We employ a simple LSTM-based \citep{hochreiterLongShortTermMemory1997} recurrent neural network (RNN) architecture applicable to a wide range of time-series problems. Physical constraints are imposed by integrating information based on the idea of physics-informed neural networks (PINNs) \citep{raissiPhysicsinformedNeuralNetworks2019}, resulting in a physics-constrained recurrent neural network (PC-RNN). The model consists of a single-layer LSTM followed by a single dense layer, reducing the dimensionality of the LSTM output to a scalar value (the temperature prediction $\hat{T}^b_t$). The detailed network architecture and hyperparameters are described in Appendix \ref{app:model_architecture}. 

For every time step $t - i$, where $i \geq 0$ is the number of lags or previous time steps observed, the network receives the state of the WTG $\mathbf{x}_{t-i}$ as input. Similar to the approach from \cite{zhangComparisonDatadrivenModelbased2014}, the network also receives information about the current time step $t$ in form of the current state $\mathbf{x}_{t}$. The network simultaneously computes the temperature gradient $\Delta T^b$ based on Equation \ref{eq:dT_neuralnetwork}. Based on \cite{gokhalePhysicsInformedNeural2022}, a simple Euler-approximation is used to get the predicted temperature gradient:
\begin{align}
\label{eq:euler_derivative}
    \Delta\hat{T}^b &= \hat{T}^b_t - T^b_{t-1}
\end{align}
The loss consists of two components, a standard prediction loss $\mathcal{L}_{pred}$ minimizing the error between the temperature prediction $\hat{T}^b_t$ and the true temperature $T^b_t$ and a physics loss $\mathcal{L}_{phys}$ minimizing the error between the neural network computed change in temperature $\Delta T^b$ based on Equation \ref{eq:dT_neuralnetwork} and the Euler-approximated temperature change $\Delta \hat{T}^b$ (Equation \ref{eq:euler_derivative}):
\begin{gather}
    \mathcal{L} = \mathcal{L}_{pred} + \alpha \mathcal{L}_{phys} \quad
    \text{with } \mathcal{L}_{pred} = \frac{1}{N} \sum_{i=1}^N (\hat{T}^b_t - T^b_t)^2, \quad
    \mathcal{L}_{phys} = \frac{1}{N} \sum_{i=1}^N (\Delta\hat{T}^b - \Delta T^b)^2
\end{gather}
The physics loss acts as a soft constraint for the PC-RNN to respect the principles of physics embedded in the equation. $\alpha$ is a scaling factor controlling the influence of the physics component on the total loss. The whole setup is shown in Figure \ref{fig:model}.

\begin{table}[b]
\vspace{-1em}
\centering
\caption{Test results for experiments with a single training WTG.}
\small
    \begin{tabular}{lccc}
Model & Plant A RMSE ($\pm$ $\sigma_{\bar{x}}$) & Plant B RMSE ($\pm$ $\sigma_{\bar{x}}$) & Plant C RMSE ($\pm$ $\sigma_{\bar{x}}$) \\
\midrule

Linear & 0.892 ($\pm$ 0.041) & 1.091 ($\pm$ 0.031) & 1.003 ($\pm$ 0.034) \\
RNN & 0.667 ($\pm$ 0.06) & \textbf{0.652 ($\pm$ 0.027)} & 0.844 ($\pm$ 0.041) \\
PCRNN & \textbf{0.636 ($\pm$ 0.033)} & 0.74 ($\pm$ 0.04) & \textbf{0.803 ($\pm$ 0.037)} \\
\end{tabular}

\label{tab:test_results}
\vspace{-.5em}
\end{table}

\section{Experiment Setup} \label{sec:experiment_setup}

The data used for the experiments consists of three datasets with 10 minutes averaged measurements originating from different wind farms (Plant A, Plant B, Plant C), covering a period from January 2022 to September 2023. See Appendix \ref{app:datasets} for a detailed description of the datasets.

We provide results for a PC-RNN with scaling parameter $\alpha = 0.25$ (fixed a-priori) and two baseline models, a simple RNN using the exact same architecture as the PC-RNN described in Section \ref{sec:pcrnn} but without the physics component and the following linear model:
\begin{align}
\label{eq:linear_model}
    T^b_{t} = \theta + \sum_{i = 1}^{M + 1} \sum_{j = 1}^N \boldsymbol{\Theta}_{i,j} \mathbf{X}_{t,i,j}
\end{align}
where $\theta$ is the intercept, $\boldsymbol{\Theta}$ are the model parameters and $\mathbf{X}_t \in \mathbb{R}^{(M+1) \times N}$ is the input data matrix at time step $t$ where the rows $i \in [1, M + 1]_{\mathbb{N}}$ are the variable vectors $((T^a_{t-k} - T^b_{t-k-1}), \omega_{t-k}, P_{t-k})$ with $k = i-1$. $M$ is the number of lags or previous time steps given to the model, $N$ is the number of variables. We use $M = 5$ for all models (PC-RNN, RNN, Linear), corresponding to a data interval of one hour. This physics-inspired model is based on \cite{zhangComparisonDatadrivenModelbased2014} and \cite{cambronBearingTemperatureMonitoring2017} (see Equation \ref{eq:dT_neuralnetwork}), who describe a similar model but without lagged states.

For each experiment, all models are sequentially trained on a subset of every wind farm dataset. 
We perform experiments with data taken from 1, 3, 6 and 9 randomly sampled turbines to assess performance with different data availability. We perform a time based train test split, using data from the whole year 2022 for training and the data from 2023 as test set for evaluation. Generalization is evaluated with data from 2023 from unseen WTGs in the training plant dataset as well as from the other unseen plant datasets. 

\section{Results} \label{sec:results}

\begin{figure}[t]
    \vspace{-.5em}
    \centering
    \includegraphics[width=\textwidth]{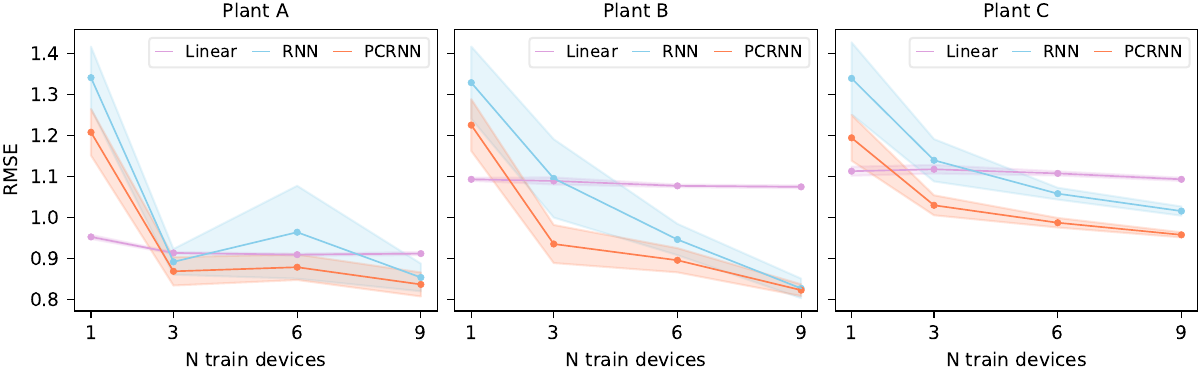}
    \caption{Generalization performance for models trained on Plant B. Error ranges $\pm 1$ SE.}
    \label{fig:results_gen}
\end{figure}

\textbf{Test performance. } Evaluations on the holdout test sets show very similar performance for the PC-RNN and standard RNN, both outperforming the linear model. When applied as a surrogate model for a single WTG - a common real-world scenario - the PC-RNN offers better performance than the conventional baseline RNN with lower error in two of the three test cases (Table \ref{tab:test_results}).

\textbf{Generalization. } The PC-RNN outperforms both baseline models in a majority of experiments. While the Linear model shows better generalization in experiments with only one training WTG, the PC-RNN performs better in the other scenarios and shows more consistent results than the baseline RNN (Figure \ref{fig:results_gen}). Generalization RMSE values for experiments with 6 sampled WTGs for training are shown in Table \ref{tab:gen_results}. Complete test and generalization results are shown in Appendix \ref{app:results}.

\begin{table}[t]
\centering
\caption{Generalization results for experiments with 6 training WTGs.}
\small
    \begin{tabular}{clccc}
Train plant & Model & Plant A RMSE ($\pm$ $\sigma_{\bar{x}}$) & Plant B RMSE ($\pm$ $\sigma_{\bar{x}}$) & Plant C RMSE ($\pm$ $\sigma_{\bar{x}}$) \\
\midrule

\multirow[c]{3}{*}{Plant A} & Linear & 0.887 ($\pm$ 0.01) & 1.078 ($\pm$ 0.002) & 1.101 ($\pm$ 0.003) \\
 & RNN & 0.784 ($\pm$ 0.017) & 1.048 ($\pm$ 0.048) & 1.032 ($\pm$ 0.015) \\
 & PCRNN & \textbf{0.726 ($\pm$ 0.013)} & \textbf{0.95 ($\pm$ 0.027)} & \textbf{0.968 ($\pm$ 0.007)} \\

\cmidrule(lr){1-1} \cmidrule(lr){2-2} \cmidrule(lr){3-3} \cmidrule(lr){4-4} \cmidrule(lr){5-5}
 \multirow[c]{3}{*}{Plant B} & Linear & 0.908 ($\pm$ 0.002) & 1.076 ($\pm$ 0.003) & 1.107 ($\pm$ 0.005) \\
 & RNN & 0.963 ($\pm$ 0.113) & 0.945 ($\pm$ 0.038) & 1.057 ($\pm$ 0.014) \\
 & PCRNN & \textbf{0.878 ($\pm$ 0.031)} & \textbf{0.895 ($\pm$ 0.03)} & \textbf{0.986 ($\pm$ 0.012)} \\
 
 \cmidrule(lr){1-1} \cmidrule(lr){2-2} \cmidrule(lr){3-3} \cmidrule(lr){4-4} \cmidrule(lr){5-5}
 \multirow[c]{3}{*}{Plant C} & Linear & 0.982 ($\pm$ 0.011) & 1.118 ($\pm$ 0.009) & 1.059 ($\pm$ 0.005) \\
 & RNN & 0.976 ($\pm$ 0.049) & 1.031 ($\pm$ 0.024) & 0.924 ($\pm$ 0.016) \\
 & PCRNN & \textbf{0.929 ($\pm$ 0.029)} & \textbf{1.028 ($\pm$ 0.02)} & \textbf{0.913 ($\pm$ 0.014)} \\
\end{tabular}

\label{tab:gen_results}
\vspace{-1em}
\end{table}

\section{Conclusion} \label{sec:conclusion}

ML-based predictive maintenance for wind energy systems can help to improve their operational efficiency and reduce costs of wind energy. Integrating physics can increase model performance, but only partially known system dynamics often impede their application. We propose a RNN architecture for predicitive maintenance of WTGs constrained by physics where unknown system coefficients are treated as trainable parameters, allowing to incorporate physics into the model when they are not or only partially known. Using nowcasting of bearing temperatures, this approach is also independent of wind speed forecasts, reducing a large factor of uncertainty. Experiments on different datasets show that the model has state of the art prediction performance and improved generalization performance compared to the baseline models, outperforming them in a majority of tests on unseen environments. 

\section*{Acknowledgements}

This work is part of the project DomLearn (892573), which has received funding in the framework of ”Energieforschung”,  a research and technology program of the Klima- und Energiefonds.

\bibliography{bibliography}

\newpage

\appendix

\section{Datasets}
\label{app:datasets}

\begin{figure}[h]
    \includegraphics[width=\textwidth]{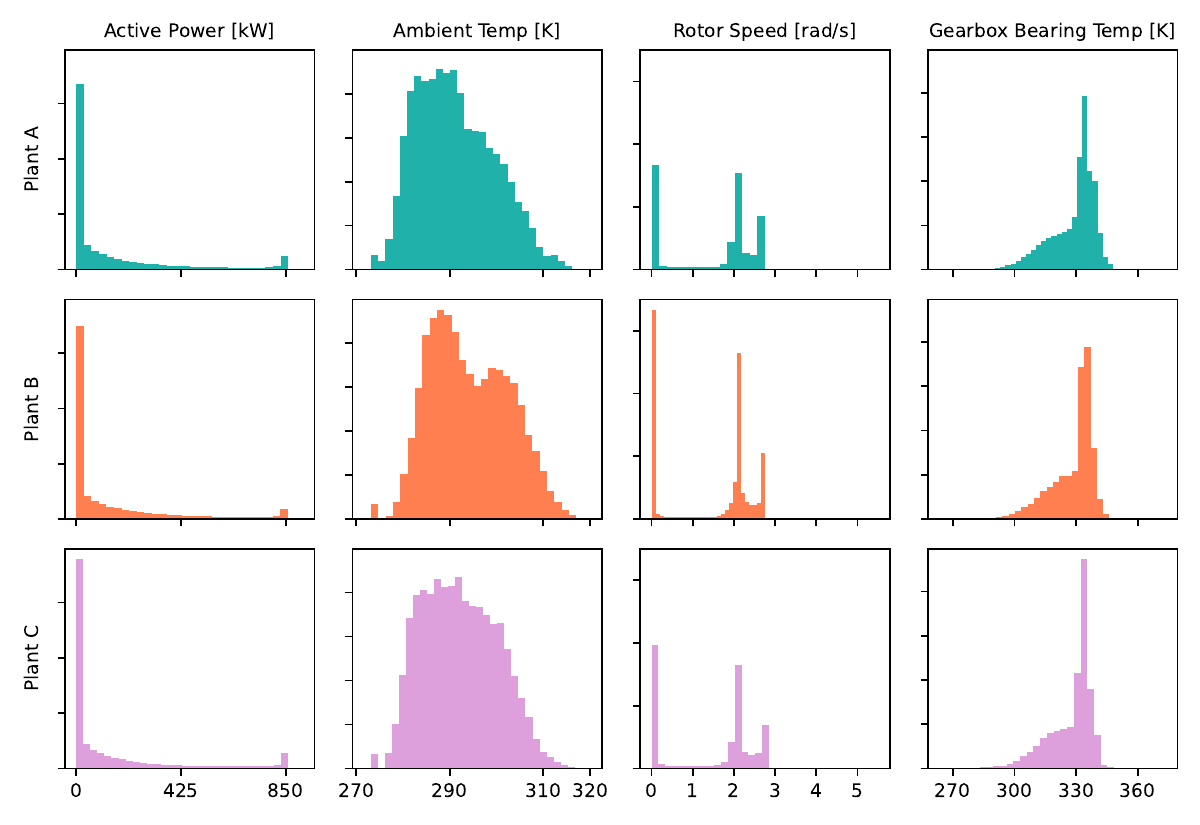}
    \caption{Feature densities for the individual datasets from different wind plants.}
    \label{fig:dataset}
\end{figure}

Table \ref{tab:dataset} shows number of individual WTGs and their maximum power output for each dataset used in the experiments. Distributions of feature values are shown in Figure \ref{fig:dataset}.
\begin{table}[h]
\centering
\caption{Number and nominal power of WTGs in each dataset.}
\small
    \begin{tabular}{lcc}
    & N WTGs & Power [kW] \\
\midrule
Plant A & 11 & 850 \\
Plant B & 50 & 850 \\
Plant C & 26 & 850 \\
\end{tabular}

\label{tab:dataset}
\vspace{-1em}
\end{table}

\section{Model architecture}
\label{app:model_architecture}
Both PC-RNN and baseline RNN consist of a single-layer LSTM cell with a hidden unit size of 16, followed by a dense layer with output size 1.
Training was performed using a batch size of 16, a 0.2 validation split and the Adam optimizer with an adaptive learning rate of 0.001. For the PC-RNN, the scaling factor $\alpha$ influencing the impact of the physics loss was set to 0.25. No exhaustive hyperparameter tuning was performed, as this was not the scope of this work.

\section{Results} \label{app:results}

\begin{figure}[t]
    \centering
    \includegraphics[width=\textwidth]{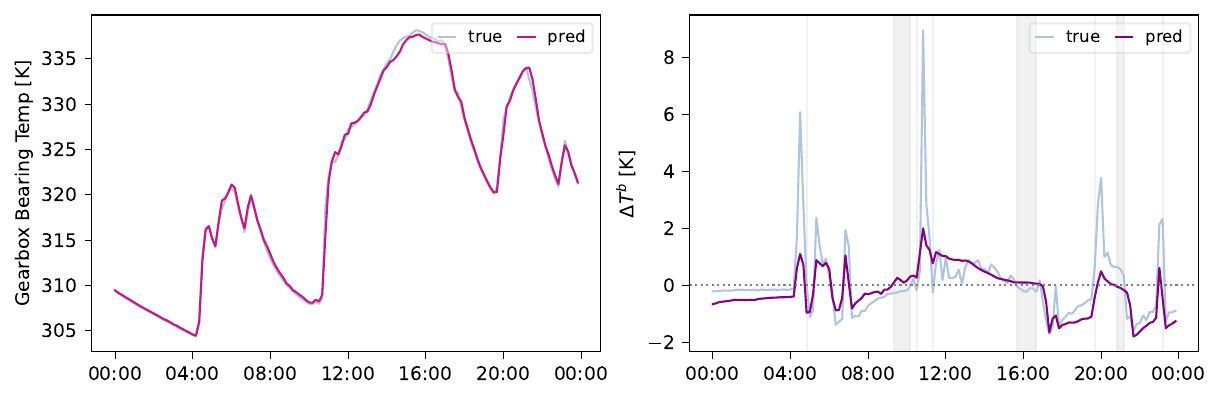}
    \caption{Example of PC-RNN $T_t^b$ prediction (left) and standardized model gradients from the PC-RNN physics component (Equation \ref{eq:dT_neuralnetwork}) with learned coefficients (right). Grey areas show time steps where model gradient and true gradient have different sign.}
    \label{fig:model_preds}
\end{figure}

Figure \ref{fig:model_preds} shows prediction results and standardized computed temperature changes from one sample PC-RNN. Solving Equation \ref{eq:dT_neuralnetwork} with learned parameters, the model is able to approximate the true pattern of temperature changes, although with smaller magnitude. Complete test results are shown in Table \ref{tab:test_results_total}.

\begin{figure}[t]
    \centering
    \includegraphics[width=\textwidth]{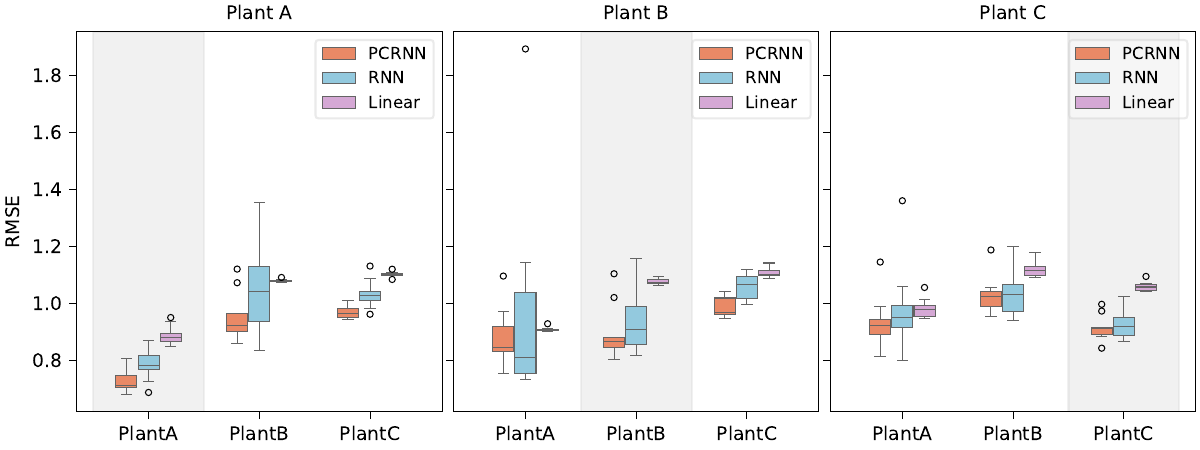}
    \caption{Model generalization performance for training set size of 6 WTGs per dataset. The grey areas mark results for unseen WTGs from training datasets (in-plant generalization).}
    \label{fig:results_gen_boxplot}
\end{figure}

Generalization performance is shown in Figure \ref{fig:results_gen_boxplot}. In most cases, the PC-RNN shows best performance and lower error variance than the standard RNN. Complete results for generalization experiments are shown in Table \ref{tab:gen_results_total}.

\begin{table}[t]
\small
\centering
\caption{Results on test set for all experiments.}
    \begin{tabular}{clcccc}
    \textbf{N train} & \textbf{Model} & \multicolumn{3}{c}{\textbf{RMSE ($\pm$ $\sigma_{\bar{x}}$)}} \\
    \cmidrule(lr){1-1} \cmidrule(lr){2-2} \cmidrule(lr){3-5}
    &  & \textbf{Plant A} & \textbf{Plant B} & \textbf{Plant C} \\
    \cmidrule(lr){3-3} \cmidrule(lr){4-4} \cmidrule(lr){5-5}

    \multirow[c]{3}{*}{1} & Linear & 0.892 ($\pm$ 0.041) & 1.091 ($\pm$ 0.031) & 1.003 ($\pm$ 0.034) \\
    & RNN & 0.667 ($\pm$ 0.06) & \textbf{0.652 ($\pm$ 0.027)} & 0.844 ($\pm$ 0.041) \\
    & PCRNN & \textbf{0.636 ($\pm$ 0.033)} & 0.74 ($\pm$ 0.04) & \textbf{0.803 ($\pm$ 0.037)} \\

    \cmidrule(lr){3-3} \cmidrule(lr){4-4} \cmidrule(lr){5-5}
    \multirow[c]{3}{*}{3} & Linear & 0.914 ($\pm$ 0.022) & 1.013 ($\pm$ 0.028) & 1.021 ($\pm$ 0.019) \\
    & RNN & 0.694 ($\pm$ 0.047) & \textbf{0.705 ($\pm$ 0.035)} & \textbf{0.796 ($\pm$ 0.025)} \\
    & PCRNN & \textbf{0.679 ($\pm$ 0.02)} & 0.722 ($\pm$ 0.025) & 0.808 ($\pm$ 0.025) \\

    \cmidrule(lr){3-3} \cmidrule(lr){4-4} \cmidrule(lr){5-5}
    \multirow[c]{3}{*}{6} & Linear & 0.914 ($\pm$ 0.01) & 1.063 ($\pm$ 0.015) & 1.012 ($\pm$ 0.008) \\
    & RNN & 0.847 ($\pm$ 0.045) & \textbf{0.722 ($\pm$ 0.017)} & \textbf{0.826 ($\pm$ 0.02)} \\
    & PCRNN & \textbf{0.768 ($\pm$ 0.05)} & 0.777 ($\pm$ 0.027) & 0.867 ($\pm$ 0.033) \\

    \cmidrule(lr){3-3} \cmidrule(lr){4-4} \cmidrule(lr){5-5}
    \multirow[c]{3}{*}{9} & Linear & 0.904 ($\pm$ 0.003) & 1.056 ($\pm$ 0.016) & 1.026 ($\pm$ 0.006) \\
    & RNN & 0.788 ($\pm$ 0.028) & 0.797 ($\pm$ 0.085) & \textbf{0.803 ($\pm$ 0.008)} \\
    & PCRNN & \textbf{0.772 ($\pm$ 0.025)} & \textbf{0.741 ($\pm$ 0.022)} & 0.839 ($\pm$ 0.018) \\
\end{tabular}
\label{tab:test_results_total}
\end{table}

\begin{landscape}
\begin{table}
\footnotesize
\centering
\caption{Generalization results for all experiments.}
    \begin{tabular}{clcccccc}
    N train & Model & \multicolumn{3}{c}{Train: Plant A} & \multicolumn{3}{c}{Train: Plant B} \\
    \cmidrule(lr){1-1} \cmidrule(lr){2-2}  \cmidrule(lr){3-5} \cmidrule(lr){6-8}
    &  & Plant A & Plant B & Plant C & Plant A & Plant B & Plant C \\
    \cmidrule(lr){3-3} \cmidrule(lr){4-4} \cmidrule(lr){5-5} \cmidrule(lr){6-6} \cmidrule(lr){7-7} \cmidrule(lr){8-8}

    \multirow[c]{3}{*}{1} & Linear & \textbf{0.937 ($\pm$ 0.005)} & \textbf{1.115 ($\pm$ 0.007)} & \textbf{1.146 ($\pm$ 0.012)} & \textbf{0.952 ($\pm$ 0.005)} & \textbf{1.092 ($\pm$ 0.004)} & \textbf{1.112 ($\pm$ 0.011)} \\
    & RNN & 1.396 ($\pm$ 0.088) & 1.666 ($\pm$ 0.111) & 1.6 ($\pm$ 0.105) & 1.34 ($\pm$ 0.076) & 1.328 ($\pm$ 0.088) & 1.338 ($\pm$ 0.088) \\
    & PCRNN & 1.232 ($\pm$ 0.071) & 1.476 ($\pm$ 0.094) & 1.424 ($\pm$ 0.091) & 1.207 ($\pm$ 0.057) & 1.224 ($\pm$ 0.063) & 1.193 ($\pm$ 0.056) \\

    \cmidrule(lr){3-3} \cmidrule(lr){4-4} \cmidrule(lr){5-5} \cmidrule(lr){6-6} \cmidrule(lr){7-7} \cmidrule(lr){8-8}
    \multirow[c]{3}{*}{3} & Linear & 0.908 ($\pm$ 0.009) & 1.082 ($\pm$ 0.006) & 1.103 ($\pm$ 0.008) & 0.913 ($\pm$ 0.002) & 1.088 ($\pm$ 0.009) & 1.117 ($\pm$ 0.011) \\
    & RNN & 1.015 ($\pm$ 0.107) & 1.29 ($\pm$ 0.092) & 1.136 ($\pm$ 0.046) & 0.891 ($\pm$ 0.031) & 1.094 ($\pm$ 0.096) & 1.139 ($\pm$ 0.052) \\
    & PCRNN & \textbf{0.855 ($\pm$ 0.04)} & \textbf{1.04 ($\pm$ 0.039)} & \textbf{1.013 ($\pm$ 0.029)} & \textbf{0.868 ($\pm$ 0.035)} & \textbf{0.934 ($\pm$ 0.046)} & \textbf{1.029 ($\pm$ 0.024)} \\

    \cmidrule(lr){3-3} \cmidrule(lr){4-4} \cmidrule(lr){5-5} \cmidrule(lr){6-6} \cmidrule(lr){7-7} \cmidrule(lr){8-8}
    \multirow[c]{3}{*}{6} & Linear & 0.887 ($\pm$ 0.01) & 1.078 ($\pm$ 0.002) & 1.101 ($\pm$ 0.003) & 0.908 ($\pm$ 0.002) & 1.076 ($\pm$ 0.003) & 1.107 ($\pm$ 0.005) \\
    & RNN & 0.784 ($\pm$ 0.017) & 1.048 ($\pm$ 0.048) & 1.032 ($\pm$ 0.015) & 0.963 ($\pm$ 0.113) & 0.945 ($\pm$ 0.038) & 1.057 ($\pm$ 0.014) \\
    & PCRNN & \textbf{0.726 ($\pm$ 0.013)} & \textbf{0.95 ($\pm$ 0.027)} & \textbf{0.968 ($\pm$ 0.007)} & \textbf{0.878 ($\pm$ 0.031)} & \textbf{0.895 ($\pm$ 0.03)} & \textbf{0.986 ($\pm$ 0.012)} \\

    \cmidrule(lr){3-3} \cmidrule(lr){4-4} \cmidrule(lr){5-5} \cmidrule(lr){6-6} \cmidrule(lr){7-7} \cmidrule(lr){8-8}
    \multirow[c]{3}{*}{9} & Linear & 0.882 ($\pm$ 0.014) & \textbf{1.077 ($\pm$ 0.001)} & 1.106 ($\pm$ 0.001) & 0.911 ($\pm$ 0.004) & 1.074 ($\pm$ 0.003) & 1.092 ($\pm$ 0.004) \\
    & RNN & 0.896 ($\pm$ 0.129) & 1.079 ($\pm$ 0.05) & 1.052 ($\pm$ 0.019) & 0.853 ($\pm$ 0.034) & 0.826 ($\pm$ 0.024) & 1.015 ($\pm$ 0.011) \\
    & PCRNN & \textbf{0.791 ($\pm$ 0.05)} & 1.086 ($\pm$ 0.076) & \textbf{1.034 ($\pm$ 0.047)} & \textbf{0.836 ($\pm$ 0.029)} & \textbf{0.822 ($\pm$ 0.014)} & \textbf{0.957 ($\pm$ 0.006)} \\

    \multicolumn{8}{c}{} \\
    \multicolumn{8}{c}{} \\
    N train & Model & \multicolumn{3}{c}{Train: Plant C} & \multicolumn{3}{c}{} \\
    \cmidrule(lr){1-1} \cmidrule(lr){2-2}  \cmidrule(lr){3-5}
    &  & Plant A & Plant B & Plant C &  &  &  \\
    \cmidrule(lr){3-3} \cmidrule(lr){4-4} \cmidrule(lr){5-5}

    \multirow[c]{3}{*}{1} & Linear & \textbf{1.006 ($\pm$ 0.028)} & \textbf{1.141 ($\pm$ 0.022)} & \textbf{1.1 ($\pm$ 0.007)} \\
    & RNN & 1.576 ($\pm$ 0.221) & 1.58 ($\pm$ 0.183) & 1.421 ($\pm$ 0.135) \\
    & PCRNN & 1.391 ($\pm$ 0.177) & 1.429 ($\pm$ 0.142) & 1.273 ($\pm$ 0.11) \\

    \cmidrule(lr){3-3} \cmidrule(lr){4-4} \cmidrule(lr){5-5} 
    \multirow[c]{3}{*}{3} & Linear & \textbf{1.014 ($\pm$ 0.019)} & 1.143 ($\pm$ 0.017) & 1.065 ($\pm$ 0.005) \\
    & RNN & 1.158 ($\pm$ 0.072) & 1.16 ($\pm$ 0.054) & 1.042 ($\pm$ 0.042) \\
    & PCRNN & 1.082 ($\pm$ 0.061) & \textbf{1.135 ($\pm$ 0.043)} & \textbf{1.003 ($\pm$ 0.029)} \\

    \cmidrule(lr){3-3} \cmidrule(lr){4-4} \cmidrule(lr){5-5} 
    \multirow[c]{3}{*}{6} & Linear & 0.982 ($\pm$ 0.011) & 1.118 ($\pm$ 0.009) & 1.059 ($\pm$ 0.005) \\
    & RNN & 0.976 ($\pm$ 0.049) & 1.031 ($\pm$ 0.024) & 0.924 ($\pm$ 0.016) \\
    & PCRNN & \textbf{0.929 ($\pm$ 0.029)} & \textbf{1.028 ($\pm$ 0.02)} & \textbf{0.913 ($\pm$ 0.014)} \\

    \cmidrule(lr){3-3} \cmidrule(lr){4-4} \cmidrule(lr){5-5}
    \multirow[c]{3}{*}{9} & Linear & \textbf{0.986 ($\pm$ 0.005)} & 1.119 ($\pm$ 0.004) & 1.055 ($\pm$ 0.004) \\
    & RNN & 0.998 ($\pm$ 0.022) & 1.031 ($\pm$ 0.013) & 0.956 ($\pm$ 0.013) \\
    & PCRNN & 1.056 ($\pm$ 0.076) & \textbf{1.023 ($\pm$ 0.012)} & \textbf{0.939 ($\pm$ 0.012)} \\

\end{tabular}
\label{tab:gen_results_total}
\end{table}
\end{landscape}

\end{document}